\documentclass{l4dc2027}

\usepackage{booktabs}
\usepackage{multirow}
\usepackage{float}
\usepackage{wrapfig}
\usepackage[dvipsnames]{xcolor}
\usepackage{titletoc}
\usepackage{verbatim}

\newcommand{\qed}{\hfill$\square$}

\newcommand{\logit}{\mathrm{logit}}

\DeclareMathOperator*{\argmax}{arg\,max}

\usepackage[font=footnotesize]{caption}

\definecolor{cobalt}{rgb}{0.02, 0.29, 0.65}

\usepackage{tikz}
\usetikzlibrary{matrix}
\usetikzlibrary{calc}
\usetikzlibrary{fit, arrows.meta, positioning}
\usetikzlibrary{decorations.pathreplacing} 

\title[Where Should a Gate Look?]{Where Should a Gate Look?\\ Input vs. State Gating in Diagonal State Space Models 
\protect\thanks{\unexpanded{
Parts of the experimental setup are adapted from an unpublished arXiv technical report by some of the authors \citep{coffee}. As discussed in the paper, the present work has a substantially different, primarily theoretical focus.
}}}
\usepackage{times}

\author{\Name{Riccardo Zattra} \Email{riccardo.zattra@phd.unipd.it}  \\
        \Name{Giacomo Baggio} \Email{baggio@dei.unipd.it} \\
        \Name{Leonardo {De Marchi}} \Email{leonardo.demarchi.2@studenti.unipd.it}  \\
        \Name{Augusto Ferrante} \Email{augusto@dei.unipd.it}  \\
        \Name{Francesco Ticozzi} \Email{ticozzi@dei.unipd.it}  \\
        \addr Department of Information Engineering\\ 
              University of Padova, Italy
        \AND
        \Name{Marco {Del Pr\`a}} \Email{marcodelpra@hotmail.it}  \\
        \addr Independent Researcher
        \AND
        \Name{Antonio Orvieto} \Email{antonio@tue.ellis.eu}  \\
        \addr Ellis Institute - Max Planck Institute for Intelligent Systems - T\"ubingen AI center
        }

\begin{document}

\maketitle

\begin{abstract}%
Modern state-space models owe much of their computational efficiency to their linear, often diagonal, dynamics. Classical recurrent neural networks instead allow nonlinear state-to-state interactions, which can be beneficial when memory updates need to depend on the past context. In this paper, we ask whether state feedback through a simple gate regulating memory updates can enhance the expressivity of diagonal state-space models. To investigate this question, we compare gates that depend on the current input, the recurrent state, or both. We establish a strict expressivity separation: for scalar models, state gating solves induction-head tasks for arbitrary vocabulary size and sequence length, whereas input gating fails on a finite instance under standard stability assumptions on the state transition matrix. Experiments show that state gating retains its advantage on induction-head tasks in higher-dimensional models, whereas input gating performs better on selective copying tasks. Together, our results show that even a simple, diagonal form of state-feedback gating can unlock capabilities unavailable to input-controlled models, while input and state gating offer complementary strengths and achieve the strongest overall~performance~when~combined.
\end{abstract}


\section{Introduction}
State-space models (SSMs) have recently emerged as a promising alternative to attention-based architectures for sequence modeling \citep{S4,smithsimplified,orvieto2023resurrecting,mamba,alonso2025state,massai2025free}. A key feature behind their computational efficiency is that their dynamics are linear in the state, often with diagonal transition matrices. In selective SSMs such as S6, the recurrent core of
Mamba \citep{mamba}, the state-update coefficients may depend on the current input, but, once the input sequence is fixed, the state evolves through a sequence of affine maps. The recurrence can therefore be evaluated sequentially with cost scaling linearly in the sequence length, or in parallel through associative scans \citep{blelloch1990prefix,martin2018parallelizing}. These properties have led to strong empirical performance in language \citep{dao2023hungry}, audio \citep{goel2022s}, and vision \citep{nguyen2022s4nd}.

Classical recurrent neural networks (RNNs) take a different approach: in particular, architectures such as Long Short-Term Memory (LSTM) \citep{hochreiter1997long} and  Gated Recurrent Units (GRUs)  \citep{cho2014properties} introduce nonlinear state-to-state interactions, in particular through \emph{gates}, i.e., variables that modulate the contribution of old and new information to the state update. These gates typically depend on both the current input and the previous state (often in a dense, nonlinear way), so that the same input can induce different updates depending on the stored information.  This additional flexibility can be useful when the appropriate update depends on the past context, and recent results show that modern RNNs can remain competitive with, and in some settings outperform, SSMs \citep{beck2024xlstm}. The price is a nonlinear recurrence that no longer admits the direct associative-scan evaluation available to input-controlled affine dynamics.

\begin{figure}[t]
    \centering
    \includegraphics[width=0.9\linewidth]{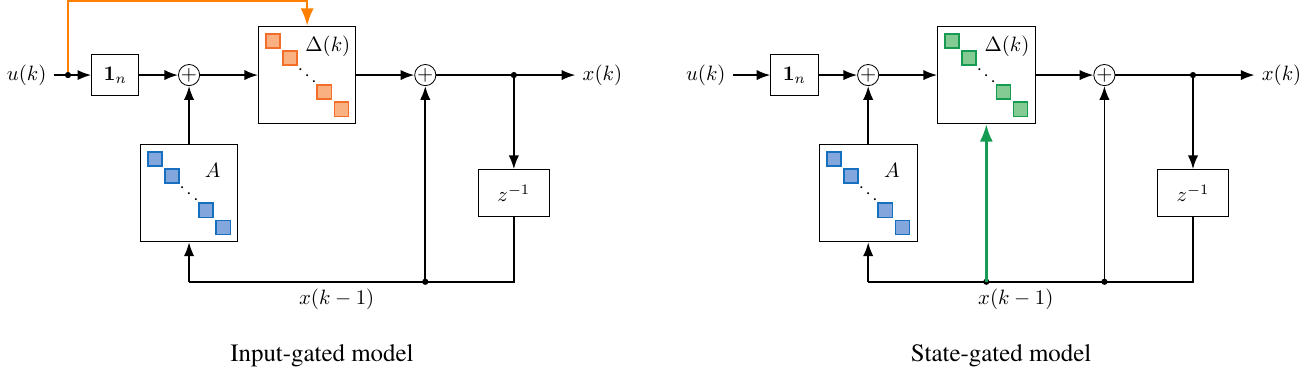}
    \caption{ Block diagrams of the input-gated (left) and state-gated (right) diagonal SSMs considered in this work. The input-gated model uses a feedforward connection from the input to compute the gate $\Delta(k)$, whereas the state-gated model uses feedback from the previous state.}
    \label{fig:gated-models}
\end{figure}

These two classes of models suggest an intermediate regime. One can introduce state feedback through a simple
gating mechanism while retaining structured dynamics that may allow efficient computation. Such a design is especially natural in deep recurrent architectures, where simple recurrent transitions can be composed across layers to improve expressivity. Motivated by this, we study a family of single-input models that share the following simple diagonal dynamics
\begin{equation}\label{eq:intro-model}
x(k)=\bigl(I+\Delta(k)A\bigr)x(k-1)+\Delta(k)\mathbf{1}_n u(k),    
\end{equation}
where the matrices $A$, $\Delta(k)\in\mathbb{R}^n$ are diagonal and $\mathbf{1}_n$ is the all-ones $n$-dimensional vector. The matrix $\Delta(k)$ plays the role of a gate: its entries lie in $(0,1)$ and control how strongly each state coordinate is affected by the recurrent dynamics and the current input.  The models differ only in how this gate is computed. As illustrated in Fig.~\ref{fig:gated-models}, in an input-gated model, the gate $\Delta(k)$ depends only on the current input $u(k)$, as in input-selective SSMs, whereas in a state-gated model it depends only on the previous state $x(k-1)$. A dual-gated model lets the gate depend on both.

For the diagonal system \eqref{eq:intro-model}, we ask:
\begin{center}
\emph{What can be gained by introducing state feedback through the gate $\Delta(k)$?}
\end{center}

We remark that the state-gated model no longer admits the direct associative scan available to its input-gated counterpart. However, its nonlinear dynamics retain a diagonal Jacobian, which can facilitate parallelization through fixed-point or (quasi) Newton iterations
based on efficient parallel scans \citep{lim2024parallelizing,gonzalez2024scalable,danieli2025pararnn,farsang2026parallelization}.

\paragraph{Contributions.}
We establish a strict expressivity separation between the input- and state-gated versions of \eqref{eq:intro-model} on the induction-head task \citep{dao2023hungry}, a paradigmatic benchmark for context-dependent memory. In particular, we show that a scalar state-gated model solves these tasks exactly for arbitrary vocabulary size and sequence length, whereas scalar input-gated models fail already for vocabulary size 3 and sequence length 5 under standard stability assumptions on the state transition matrix. The same impossibility result applies in particular to scalar S6.
Thus, state feedback through gating can overcome a structural limitation of input-controlled dynamics, even in the scalar setting. Our experiments investigate whether this advantage persists in higher dimensions and extend the comparison to other benchmarks, including selective copying \citep{mamba} and MNIST classification. State-dependent gating substantially outperforms input gating on induction-head tasks, whereas selective copying favors input gating because the decision of whether to store or ignore a token depends largely on the current token itself. Together, our results show that a simple form of state-feedback gating can provably enhance expressivity on certain tasks, with a diagonal structure that may enable efficient implementation. At the same time, input and state gating offer distinct advantages across tasks, with their combination yielding the~strongest~overall~performance.

\section{Related work}
\label{sec:related_work}

\paragraph{Expressivity of recurrent models and SSMs.} The expressivity of recurrent sequence models has been studied from several perspectives. Classical results show that (nonlinear) RNNs can approximate broad classes of dynamical input-output maps over finite time horizons \citep{funahashi1993approximation,schafer2007recurrent}, with extensions to infinite horizons \citep{hanson2020universal} and deep recurrent architectures, including LSTMs and GRUs \citep{song2023minimal}. SSMs (or linear RNNs) can approximate broad classes of convolution kernels with sufficient width \citep{li2022linear}, while combining them with nonlinear projections or readouts can yield universal approximation of sequence-to-sequence maps \citep{orvieto2023universality,wang2023state}. For selective SSMs, \cite{cirone2024theoretical} characterize broad classes of input-controlled linear systems, while \cite{huang2025understanding} analyze how input selectivity in S6 affects its approximation power. 
Complementary work investigates the formal computational power of RNNs \citep{siegelmann1995computational,weiss2018practical,merrill2020formal} and the limitations of SSM-based models on state tracking and symbolic computation \citep{merrill2024illusion,shakerinava2026expressive, terzic2026structured}. 
For copying and retrieval tasks, \citet{jelassi2024repeat} show that any recurrent model with finite-state memory requires memory growing linearly with sequence length to copy entire input sequences, while \citet{arora2024zoology} show that models combining linear filtering and input-dependent gating are less effective than attention at retrieving multiple previously observed associations.

Here, we study how the source of the gating signal affects the expressive power of gated diagonal SSMs, establishing a strict separation between input- and state-dependent gating on induction-head tasks. This complements the results of \citet{huang2025understanding}, who show that input-controlled models can solve induction-head tasks with additional width and architectural structure, whereas our comparison keeps the recurrence and state dimension fixed. While our theoretical results establish an expressive advantage of state feedback on induction-head tasks, this advantage is task-dependent, as our selective-copying experiments show that input gating can be more effective.

\paragraph{State dependence in recurrent sequence models.} State-dependent memory updates are a classical feature of gated recurrent architectures such as LSTMs and GRUs, and remain central in more recent models such as xLSTM \citep{beck2024xlstm}. In these models, the previous state influences the gates, making the dynamics nonlinear. Conversely, \citet{feng2024were} show that removing state dependence from LSTM and GRU gates yields simplified architectures amenable to parallel training. Other sequence models use the recurrent state differently. DeltaNet, Gated DeltaNet \citep{yang2024parallelizing,yang2025gated}, RWKV-7 \citep{peng2025rwkv}, and Comba \citep{hu2026improving} incorporate state-dependent memory corrections while retaining affine state updates for fixed inputs, allowing scan-based parallelization. Other architectures, including TTT, Titans, and MIRAS \citep{sun2025ttt,behrouz2024titans,behrouz2025s}, instead employ nonlinear state-dependent memory dynamics, for instance through online optimization or nonlinear memory modules. 

Here, we focus on a simpler form of state feedback, acting only through a component-wise gate and preserving a diagonal transition Jacobian. Related structured nonlinear dynamics have been considered in \citep{danieli2025pararnn,farsang2026parallelization}, where structured Jacobians are exploited for efficient parallelization. In particular, our state-gated model without the gate bias coincides with COFFEE \citep{coffee}. The latter work introduced this recurrence and evaluated its empirical performance across several tasks. We extend COFFEE by introducing a learnable gate bias and, more importantly, proving an expressivity separation between state and input gating. We also study a dual-gated variant combining both signals.

\section{A family of gated diagonal SSMs}
\label{sec:COFFEE}

In order to study how the source of the gating signal affects the behavior and expressive power of a diagonal state-space model, we introduce three matched SSMs that share the same recurrence and differ only in whether the gate is computed from the current input, the recurrent state, or both. We refer to these models as input-gated (IG), state-gated (SG), and dual-gated (DG), respectively. As detailed in Appendix \ref{app:s6_connection}, IG is a simplified version of the S6 layer in Mamba \citep{mamba}, retaining its input-dependent gating mechanism while simplifying the discretization and removing other input-dependent components that would obscure the role of the gate.

We stress that our object of study is the selective recurrence itself, not the surrounding architecture in which it is typically embedded. SSM-based blocks such as Mamba often incorporate convolutional or other input-processing modules around the core recurrence, which can provide some degree of input-side contextualization independently of the gate. To ensure a fair comparison of gating mechanisms in SSMs, we study the SSM layer in isolation, without any surrounding blocks (e.g. convolutions, MLPs etc.) that could~confound~its~behavior.

Let $u(k)\in\mathbb{R}^D$ denote the input embedding at time $k$, and let $u_i(k)$ be its $i$-th component. Each input feature is processed by a separate state-space subsystem with state $x^{(i)}(k)\in\mathbb{R}^n$:
\begin{equation}
\label{eq:shared_gated_recurrence}
\left\{\begin{aligned}
x^{(i)}(k) &= (I+\Delta^{(i)}(k)A^{(i)})x^{(i)}(k-1)+\Delta^{(i)}(k)\mathbf{1}_n u_i(k), \\
y^{(i)}(k) &= C^{(i)}x^{(i)}(k),
\end{aligned}\right.\qquad i=1,\ldots,D
\end{equation}
where $A^{(i)}\in\mathbb{R}^{n\times n}$ is diagonal, $C^{(i)}\in\mathbb{R}^{1\times n}$ a learnable readout, and $\mathbf{1}_n\in\mathbb{R}^n$ the~all-ones~vector.

The matrix $\Delta^{(i)}(k)\in\mathbb{R}^{n\times n}$ acts as a component-wise gate and is taken to be diagonal:
$$
\Delta^{(i)}(k)=\operatorname{diag}(\delta^{(i)}_1(k),\ldots,\delta^{(i)}_n(k)),\qquad\delta^{(i)}_j(k)\in(0,1).
$$
Each coefficient $\delta^{(i)}_j(k)$ controls, for the corresponding state coordinate, the balance between retaining the previous state and incorporating the current input.

In all three models introduced below, the gating coefficients \smash{$\delta_j^{(i)}(k)$} are obtained by applying the sigmoid function component-wise to the corresponding gating signal, ensuring that \smash{$\delta_j^{(i)}(k)\in(0,1)$}. For numerical stability, the diagonal entries of $A^{(i)}$ are constrained so that $I+\Delta^{(i)}(k)A^{(i)}$ remains Schur stable for every admissible
gate.  

\paragraph{Input-gated model.} The input-gated model computes the gate from the current input feature:

\begin{equation}
\label{eq:ig_gate}
\Delta_{\mathrm{IG}}^{(i)}(k)=\operatorname{diag}\left(\sigma\left(w_u^{(i)}u_i(k) + b^{(i)}\right)\right),\qquad w_u^{(i)},b^{(i)}\in\mathbb{R}^n,
\end{equation}
where $\sigma(\cdot)$ denotes the sigmoid function and is applied component-wise.
In IG, fixing the current input also fixes the gate and, consequently, the coefficients of the affine state update. The recurrent state still stores information about the past, but it does not directly influence how the current update is gated. IG therefore implements input-based, or token-based, selectivity: the same input, when it appears at different time steps, produces the same gating pattern regardless of the preceding context. 
\paragraph{State-gated model.} The state-gated model computes the gate from the state reached before the current input:
\begin{equation}
\label{eq:sg_gate}
\Delta_{\mathrm{SG}}^{(i)}(k)=\operatorname{diag}\left(\sigma\left(w_x^{(i)}\odot x^{(i)}(k-1) + b^{(i)}\right)\right),\qquad w_x^{(i)},b^{(i)}\in\mathbb{R}^n,
\end{equation}
where $\odot$ denotes the Hadamard product. Because the state summarizes the previously processed sequence, SG can adapt its update to the accumulated context. In particular, the same input (i.e. $u(k)$ for $k \in {1,\ldots,L}$) can induce different gates 
depending on the history that precedes it. SG therefore implements state-based, or context-based, selectivity. The dependence of the gate on the state makes the recurrence nonlinear. However, since both $A^{(i)}$ and $\Delta_{\mathrm{SG}}^{(i)}(k)$ are diagonal, each state coordinate depends only on its own previous value and on the current input feature. The resulting transition Jacobian is diagonal, a structure that can reduce the cost of parallelizing the nonlinear recurrence using fixed-point or quasi-Newton methods \citep{danieli2025pararnn,farsang2026parallelization,gonzalez2024scalable}.

\paragraph{Dual-gated model.} IG and SG represent the two extremes of the design space: in IG, the gate depends only on the current input, whereas in SG it depends only on the past state. The dual-gated model combines these two sources of information within the same gate:
\begin{equation}
\label{eq:dg_gate}
\Delta_{\mathrm{DG}}^{(i)}(k)=\operatorname{diag}\left(\sigma\left(w_u^{(i)}u_i(k)+w_x^{(i)}\odot x^{(i)}(k-1) + b^{(i)}\right)\right),\ \ w_u^{(i)},w_x^{(i)}, b^{(i)}\in\mathbb{R}^n,
\end{equation}
The weights $w_u^{(i)}$ and $w_x^{(i)}$ allow each gate coordinate to respond primarily to the current input, the accumulated state, or both. DG includes IG and SG as special cases, recovered by setting $w_x^{(i)}=0$ and $w_u^{(i)}=0$, respectively. As in SG, state-dependent gating makes the recurrence nonlinear, while its component-wise structure keeps the state-transition Jacobian diagonal.

\begin{remark}[Parameter counts]
Each diagonal matrix $A^{(i)}$, readout $C^{(i)}$, and gating weight or bias vector contributes $n$ learnable parameters. Thus, IG and SG have $4n$ parameters per feature, while DG has $5n$. Without gate biases, these counts reduce to $3n$ and $4n$, respectively. Since each of the $D$ input features has its own subsystem, the total parameter counts are multiplied by $D$.
\end{remark}

Next we analyze how the nature 
of the gate affects model expressivity: Section~\ref{sec:analytical_results} establishes an expressivity separation between IG and SG, while Section~\ref{sec:experiments} evaluates all three models empirically.

\section{A formal expressivity separation}
\label{sec:analytical_results}

\begin{figure}[t]
    \centering
    \includegraphics[width=0.9\linewidth]{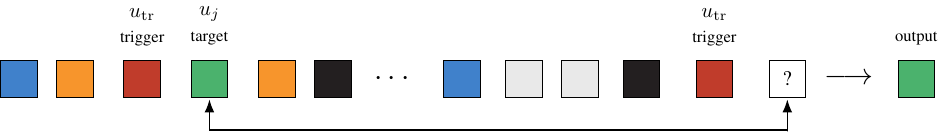}
    \caption{Illustration of the induction-head task. The trigger token (red) appears twice, and the model must recall the target token (green), which immediately follows the first trigger, when the trigger appears again. Other colors represent tokens distinct from both the trigger and the target.}
    \label{fig:induction-head}
\end{figure}

We ask whether the source of the gating signal induces a genuine expressivity separation, beyond differences in empirical performance. To investigate this question, we consider the induction-head task, a simple yet paradigmatic benchmark for evaluating the memory capabilities of sequence models, also used to evaluate Mamba \citep{dao2023hungry,mamba}. The task requires a recurrent model to store the token following a designated trigger and retrieve it when the trigger appears again, possibly after processing many other tokens. We show that this task reveals a strict expressivity separation between input- and state-gated models. More precisely, we consider the induction-head task of \citet{dao2023hungry} with triggers and targets of unit length. Let $V=\{u_1,\ldots,u_{d_V}\}$ be a vocabulary of $d_V$ scalar token embeddings, and let $L$ denote the sequence length. A designated trigger token $u_{\rm tr}\in V$ appears exactly twice in each admissible sequence: first at an arbitrary position, followed by a non-trigger token, and then at the final position. If the first trigger is followed by $u_j\in V\setminus\{u_{\rm tr}\}$, the required output after the final trigger is $u_j$. All other positions may contain arbitrary non-trigger tokens. Fig.~\ref{fig:induction-head} illustrates the task. We say that a scalar model \emph{solves} the task if there exist token embeddings, an initial condition, and a nearest-neighbor readout over $V$ such that the prediction is correct for every admissible sequence. We first show that state-dependent gating allows even a scalar SSM to solve induction-head tasks of arbitrary~vocabulary~size~and~sequence~length.

\begin{theorem}[Scalar SG can solve induction-head tasks] \label{thm:thm1}
For every vocabulary size $d_V\geq 2$ and every sequence length $L\geq 3$, there exist scalar embeddings $V=\{u_1,\ldots,u_{d_V}\}\subset\mathbb{R}$
and an initial condition $x(0)\in\mathbb{R}$ such that the SG model \eqref{eq:shared_gated_recurrence}-\eqref{eq:sg_gate}, with $D=n=1$, parameters
$$
A^{(1)}=-1,\qquad w_x^{(1)}=1,\qquad b^{(1)}=0,
$$
and readout $y(k)=x(k)$, solves with perfect accuracy the induction-head task.
\end{theorem}

The proof, given in Appendix~\ref{app:proof_thm1}, is constructive. We choose embeddings and an initial condition such that the first trigger drives the state to a large positive value, opening the gate to store the subsequent target token. The gate then nearly closes, preserving the target until the end of the sequence. We show that the state remains sufficiently close to the target embedding to guarantee correct nearest-neighbor decoding for every admissible sequence. 

We remark that Theorem~\ref{thm:thm1} is an expressivity result. It establishes the existence of real-valued embeddings, initial conditions, and model parameters for which the recurrence achieves perfect accuracy on the task. It does not address whether such a configuration can be found by optimization, nor its robustness under finite-precision numerical implementations. 

The next result shows that scalar models whose gates depend only on the current input token are not sufficient to solve even very short instances of the induction-head tasks.
\begin{theorem}[Scalar input gating fails on some induction-head tasks] \label{thm:thm2}
There exists an induction-head task with vocabulary size $d_V=3$ and sequence length $L=5$ that cannot be solved with perfect accuracy by any scalar model of the form
$$
x(k)=a(u(k))x(k-1)+b(u(k)),\qquad -1 < a(u(k))<1,
$$
where the coefficients $a(u(k))$ and $b(u(k))$ depend only on the current input token $u(k)$ and the decision is based on a threshold applied to any strictly monotone function of the final state $x(L)$.
\end{theorem}

The proof, given in Appendix~\ref{app:proof_thm2}, exploits the affine structure of the input-dependent dynamics to derive weighted-average identities between final states associated with different targets, ruling out exact recovery through nearest-neighbor decoding. The result applies, in particular, to scalar IG and S6 models with Schur-stable state transition coefficients (i.e., $|a(u)|<1$ for every input token),
but extends beyond these specific models. Indeed,
(i) it provides a no-go result for \emph{every} choice of the functions $a(\cdot)$ and $b(\cdot)$
satisfying the stated conditions: the separation therefore depends on the nature of the gating (input- versus state-dependent) rather than on a specific gating type,
(ii) the conditions on $a(\cdot)$ are more general than those of S6, since $a(\cdot)$ may take
negative values. This is relevant in view of \citep{grazzi2025unlocking}, where it is shown that
negative values improve the expressivity of linear RNNs on certain tasks,  and
(iii) the result holds for any (possibly
nonlinear) readout that is  monotone in the state,  not only for linear or affine readouts.



\section{Empirical validation}
\label{sec:experiments}

Theorems~\ref{thm:thm1} and~\ref{thm:thm2} establish a strict expressivity separation in the scalar setting: state gating can solve induction-head tasks that input gating provably cannot. We now investigate whether this advantage persists in higher-dimensional models trained end-to-end on different tasks. We also investigate whether the dual-gated model \eqref{eq:shared_gated_recurrence}-\eqref{eq:dg_gate}, which combines input and state information, offers additional advantages over the two pure-gating models. We consider two complementary synthetic benchmarks. The first is \emph{induction head} \citep{dao2023hungry}, analyzed in the previous section, which allows us to test whether the expressivity advantage of state gating persists in higher dimensions. The second is \emph{selective copying} \citep{mamba}, another important benchmark for sequence models, where the decision to store or ignore a token depends primarily on the token itself, potentially favoring input gating. Together, these tasks allow us
to investigate the complementary strengths of input and state gating and whether DG can combine them effectively. We additionally consider MNIST classification as a non-synthetic benchmark, with no clear a priori preference for either gating mechanism, to assess whether these findings extend beyond synthetic tasks.



\paragraph{Experimental setup.}
We compare the IG, SG, and DG models introduced in Section~\ref{sec:COFFEE}, which share a common core architecture so that differences in performance can be attributed to the gating mechanism. Where relevant, we also include S6, implemented as an isolated recurrent core within the same surrounding architecture used for IG, SG, and DG (Appendix \ref{appendix:core_module_architecture}), as an established input-selective baseline. This ensures the comparison remains at the level of the recurrence alone, rather than conflating it with the additional components of the full Mamba block. Each configuration is evaluated over three random seeds, and results are reported as mean $\pm$ standard deviation. For each task, we train with every learning rate (lr) in the half-decade
logarithmic grid $\{10^{-4},\allowbreak 3{\cdot}10^{-4},\allowbreak10^{-3},3{\cdot}10^{-3},10^{-2},3{\cdot}10^{-2},10^{-1}\}$
and select the value with the highest mean validation accuracy over the three
seeds for the majority of the models. We report the test accuracy at the selected learning rate, together
with the value itself in the tables. 
Moreover, we ask what changes when the learnable bias $b^{(i)}$ in~(\ref{eq:ig_gate})-(\ref{eq:dg_gate}) is removed from the gate of our matched family.
Architecture, initialization, reproducibility, and hardware details are provided in Appendix~\ref{app:exp_details}.\footnote{Our code is available at \url{https://github.com/RiccardoZattra/Where-Should-a-Gate-Look}}

\paragraph{Induction Head.}
For this experiment, we adopt the same induction head task of \cite{dao2023hungry} described in Section \ref{sec:analytical_results}, with embedding dimension $D=64$, sequence length $L_{\mathrm{seq}}=128$, and vocabulary size $|V|=1024$. All models are trained for $50$ epochs, using $10000$ batches per epoch with batch sizes $64/256/256$ (for training, validation and test). Accuracy is computed per token: an instance counts as correct only if the model outputs the target token at the end of its response. Further details are provided in Appendix~\ref{app:exp_details}. Note that, IG and SG have the same number of learnable parameters, whereas DG uses additional parameters to combine input and state information. As shown in Table~\ref{tab:ih_added_new}, IG fails to learn the task, while SG attains substantially higher accuracy, consistent with the scalar separation carrying over to the higher-dimensional trained setting. S6 also fails, despite its considerably larger parameter count. DG achieves the best results, reaching approximately $99\%$ accuracy in all runs.






\begin{table}[H]
\centering
\renewcommand{\arraystretch}{0.5}
\setlength{\tabcolsep}{5pt}
\begin{tabular}{@{} l c c c c c c c @{}}
\toprule
& & \multicolumn{3}{c}{\textbf{without bias}} 
& \multicolumn{3}{c}{\textbf{with bias}} \\
\cmidrule(lr){3-5}\cmidrule(lr){6-8}
\textbf{Model} & $n$ 
& \textbf{SSM par.} & \textbf{lr} & \textbf{Test Accuracy}
& \textbf{SSM par.} & \textbf{lr} & \textbf{Test Accuracy} \\
\midrule
IG & 8 & 1,536 & $3{\cdot}10^{-3}$ & $0.53 \pm 0.23$\% 
& 2,048 & $3{\cdot}10^{-3}$ & $0.53 \pm 0.46$\% \\

SG & 8 & 1,536 & $3{\cdot}10^{-3}$ & $83.6 \pm 5.0$\% 
& 2,048 & $3{\cdot}10^{-3}$ & $32.7 \pm 38.6$\% \\

DG & 8 & 2,048 & $3{\cdot}10^{-3}$ & $99.6 \pm 0.4$\% 
& 2,560 & $3{\cdot}10^{-3}$ & $99.5 \pm 0.6$\% \\

\midrule
\multicolumn{8}{c}{S6 ($n=8$): SSM par.\ 14{,}784, Test Accuracy $2.0 \pm 0.4$\%} \\
\bottomrule
\end{tabular}
\caption{Induction-head results (mean $\pm$ standard deviation over three seeds) at fixed state dimension $n=8$. IG, SG, and DG form the main gating comparison; S6 is included as an established input-selective baseline.}
\label{tab:ih_added_new}
\end{table}

The lower accuracy of SG with a gate bias on induction head appears to be an optimization issue rather than an expressivity limitation. Indeed, Appendix~\ref{app:bias_warm_start} shows that when training is resumed from an SG model trained without a bias, introducing a learnable bias initialized to zero improves accuracy from $32.7 \pm 38.6$\% to $99.4 \pm 0.5$\%. This suggests that the additional bias makes it more difficult to find a good solution when training from scratch. 

These experiments extend the scalar theoretical separation result. Successful learning is consistently associated with state-dependent gating: purely input-gated models consistently fail, whereas both SG and DG achieve high accuracy. DG further improves upon SG, suggesting that input information may be beneficial when combined with state feedback.

\paragraph{Selective Copying.} 
We next consider the selective-copying task as in~\citep{mamba}. A sequence contains a small number of content tokens interspersed with blank tokens, followed by a block of marker tokens. The model must reproduce the content tokens, in their original order, at the marker positions while ignoring the blanks. Unlike induction head, the decision of whether the current token should be stored or ignored depends primarily on the token itself rather than on its relationship with the preceding context, making this task a natural benchmark for input-dependent gating. We follow the standard configuration of~\citep{mamba}, using alphabet size $A=10$, noise length $L_{\mathrm{noise}}=100$ (the number of blank tokens), embedding dimension $D=64$, and state dimension $n=8$. To assess how the models behave as the task becomes more challenging, we vary the number of content tokens to be recalled, $M\in\{2,5\}$. Here, accuracy is computed per sequence, not per token: a sequence counts as correct only if all $M$ content tokens are reproduced correctly at the marker positions. The architecture and optimization settings are as follows: an isolated SSM with $D=64$, $n=8$, trained for $50$ epochs, $10000$ batches per epoch, batch sizes $128/256/256$ for training, validation and test. For more details see Appendix~\ref{app:exp_details}. 
Table~\ref{tab:sc_results_new} reports the results. For the easiest setting ($M=2$), all models solve the task almost perfectly. As the difficulty increases, however, clear differences emerge. At $M=5$, IG outperforms SG, consistent with the intuition that input-dependent gating is better suited to deciding which tokens should be stored. Nevertheless, the best performance is achieved by DG, which substantially exceeds both pure-gating models despite using only a modest increase in the number of parameters.

\begin{table}[H]
\centering
\renewcommand{\arraystretch}{0.5}
\setlength{\tabcolsep}{5pt}
\begin{tabular}{@{} c l c c c c c c @{}}
\toprule
& & \multicolumn{3}{c}{\textbf{without bias}} & \multicolumn{3}{c}{\textbf{with bias}} \\
\cmidrule(lr){3-5}\cmidrule(lr){6-8}
$M$ & \textbf{Model} & \textbf{SSM par.} & \textbf{lr} & \textbf{Test Accuracy}
 & \textbf{SSM par.} & \textbf{lr} & \textbf{Test Accuracy} \\
\midrule
\multirow{3}{*}{2}
 & IG & 1{,}536 & $3{\cdot}10^{-3}$ & $100.0 \pm 0.0$\% & 2{,}048 & $3{\cdot}10^{-3}$ & $100.0 \pm 0.0$\% \\
 & SG & 1{,}536 & $3{\cdot}10^{-3}$ & $95.9 \pm 5.3$\% & 2{,}048 & $3{\cdot}10^{-3}$ & $99.7 \pm 0.2$\% \\
 & DG & 2{,}048 & $3{\cdot}10^{-3}$ & $100.0 \pm 0.0$\% & 2{,}560 & $3{\cdot}10^{-3}$ & $100.0 \pm 0.0$\% \\
 \midrule
\multicolumn{8}{c}{S6 ($M=2$): SSM par.\ 14{,}784, lr $3{\cdot}10^{-3}$, Test Accuracy $100.0 \pm 0.0$\% } \\
\midrule
\multirow{3}{*}{5}
 & IG & 1{,}536 & $3{\cdot}10^{-3}$ & $26.7 \pm 19.2$\% & 2{,}048 & $3{\cdot}10^{-3}$ & $76.5 \pm 22.3$\% \\
 & SG & 1{,}536 & $3{\cdot}10^{-3}$ & $9.0 \pm 1.4$\% & 2{,}048 & $3{\cdot}10^{-3}$ & $27.5 \pm 9.2$\% \\
 & DG & 2{,}048 & $3{\cdot}10^{-3}$ & $88.1 \pm 2.2$\% & 2{,}560 & $3{\cdot}10^{-3}$ & $97.2 \pm 2.7$\% \\
\midrule
\multicolumn{8}{c}{S6 ($M=5$): SSM par.\ 14{,}784, lr $3{\cdot}10^{-3}$, Test Accuracy $64.8 \pm 11.0$\% } \\
\bottomrule
\end{tabular}
\caption{Selective copying across target lengths $M$ (with $A=10$,
$L_{\mathrm{noise}}=100$), mean $\pm$ standard deviation over three seeds. IG, SG, and DG form the main gating comparison; S6 is included as an established input-selective baseline.}
\label{tab:sc_results_new}
\end{table}

Taken together with the induction-head results, these experiments support a complementary picture of the two gating mechanisms. State-dependent gating appears essential for induction head, whereas input-dependent gating becomes comparatively more effective on selective copying. Across these two tasks, DG consistently achieves the strongest performance among the models we test, suggesting that jointly exploiting input and state information can provide a more robust gating strategy than relying on either signal alone.

\paragraph{MNIST.} 
Finally, we consider a sequential formulation of MNIST \citep{lecun1998gradient} as a non-synthetic benchmark, in order to verify that the previously proposed gating strategies remain effective on real-world data.
Unlike the synthetic tasks, which are designed to investigate specific gating mechanisms, MNIST provides a complementary assessment of the different gating strategies on a qualitatively different problem. We use the architecture described in Appendix~\ref{appendix:mnist_architecture}, which processes each image as sequences of rows and columns in both forward and reverse directions. Each $28\times 28$ 
image is processed as a sequence of $28$ tokens, one for each image row. All models are trained for $50$ epochs and batch size $256$. As shown in Table~\ref{tab:mnist_n2_new}, IG, SG, and DG achieve comparable test accuracies, with differences that are small relative to the variability across runs. In contrast, the standalone S6 model performs substantially worse.\footnote{The reported results refer exclusively to the standalone S6 model. As noted earlier, the goal of this work is solely to compare gating strategies within SSMs.} 

Overall, these results suggest that the choice of gating mechanism has little impact on classification performance, with state-dependent and dual gating achieving accuracy comparable to input gating despite different memory-update mechanisms. At the same time, the substantial performance gap between IG and S6 suggests that the specific parametrization of input-dependent gating can strongly affect performance.
\begin{table}[H]
\centering
\renewcommand{\arraystretch}{0.5}
\setlength{\tabcolsep}{5pt}
\begin{tabular}{@{} l c c c c c c c @{}}
\toprule
& & \multicolumn{3}{c}{\textbf{without bias}} & \multicolumn{3}{c}{\textbf{with bias}} \\
\cmidrule(lr){3-5}\cmidrule(lr){6-8}
\textbf{Model} & $n$ & \textbf{SSM par.} & \textbf{lr} & \textbf{Test Accuracy}
 & \textbf{SSM par.} & \textbf{lr} & \textbf{Test Accuracy} \\
\midrule
IG & 2 & $672$ & $10^{-2}$ & $95.2 \pm 1.6$\% & $896$ & $10^{-2}$ & $96.2 \pm 0.6$\% \\
SG & 2 & $672$ & $10^{-2}$ & $95.2 \pm 1.2$\% & $896$ & $10^{-2}$ & $95.8 \pm 0.8$\% \\
DG & 2 & $896$ & $10^{-2}$ & $95.6 \pm 0.8$\% & $1{,}120$ & $10^{-2}$ & $96.1 \pm 1.1$\% \\
\midrule
\multicolumn{8}{c}{S6 ($n=2$): SSM par.\ $11{,}312$, lr $10^{-2}$, Test Accuracy $14.3 \pm 1.8$\%} \\
\bottomrule
\end{tabular}
\caption{MNIST results. Results are reported as mean $\pm$ std over three seeds. All models (IG, SG, DG, and S6) are evaluated using the same architecture. See Appendix~\ref{appendix:mnist_architecture} for details.}
\label{tab:mnist_n2_new}
\end{table}


\section{Conclusions}
\label{sec:conclusions}

We investigated whether state feedback, introduced via a simple gating mechanism, enhances the expressivity of diagonal state-space models. Unlike input gating, state gating makes the update coefficients depend on the stored state, yielding nonlinear state-to-state interactions while preserving a diagonal structure. In the scalar setting, we prove a strict expressivity separation: state gating solves induction-head tasks exactly, whereas input-gated models, including scalar S6, provably fail on a finite instance. Experiments confirm that this advantage persists in higher-dimensional models trained end to end, while selective copying reveals a complementary regime favoring input gating. A dual-gated variant combining both signals performs best or remains competitive across tasks. Overall, state feedback overcomes fundamental limitations of input-controlled dynamics, and dual gating exploits the strengths of both mechanisms.

Our work also leaves open several questions and suggests directions for future research. First, state-dependent gating sacrifices the direct associative-scan parallelism of input-gated models. However, the resulting nonlinear dynamics retain a diagonal Jacobian, which may facilitate iterative parallelization \citep{gonzalez2024scalable, danieli2025pararnn, farsang2026parallelization}. Developing scalable implementations and assessing their practical efficiency remain important directions for future work. Second, we focus on diagonal state transitions, showing how state feedback improves selectivity by letting the recurrent state control which information is stored or retained. More demanding state-tracking tasks, however, may benefit from richer, non-diagonal transitions \citep{merrill2024illusion}. Combining state-dependent gating with structured non-diagonal transitions, such as Householder/WY parameterizations or Gated DeltaNet-style constructions \citep{yang2025gated, siems2026deltaproduct}, is therefore a natural direction to explore. Finally, our theory is restricted to scalar models with bounded gates and stable state transitions, and our experiments focus on synthetic benchmarks and MNIST. Extending the theory to higher-dimensional models and testing these mechanisms on realistic sequence-modeling tasks would clarify the broader implications of our findings.

\subsection*{Generative AI Use}
Large language models were used to assist with manuscript editing and improving clarity. All AI-assisted content was reviewed by the authors, who take full responsibility for the content and correctness of the paper.

\bibliography{L4DC/bibliography}

\clearpage

\appendix

\startcontents[appendices]
\printcontents[appendices]{l}{1}{
    \noindent\textbf{\large Appendix: Table of Contents}
    \par\smallskip
    \setcounter{tocdepth}{2}
}

\section{Connection between IG and S6}
\label{app:s6_connection}

This appendix clarifies the connection between the input-gated (IG) model introduced in Section \ref{sec:COFFEE} and the S6 layer of Mamba
\citep{mamba}. We start from a version of S6 in which the input-dependent discretization step is computed through a full linear projection, rather than the low-rank parameterization used in Mamba. This provides a convenient starting point for relating the two models. We then simplify the discretization and remove additional input-dependent components to isolate the
gating mechanism. Finally, we introduce the component-wise gate used in IG and show that the input-to-state vector can be normalized to $\mathbf{1}_n$ without loss of expressivity. The normalization argument also applies to SG and DG.

\subsection{From S6 to the IG model}

Let $u(k)\in\mathbb{R}^D$ denote the input embedding
at time $k$, with $u_i(k)$ its $i$-th component.
The S6 recurrence for each input feature takes the form
\begin{equation*}
\begin{aligned}
x^{(i)}(k) &=e^{A^{(i)}\Delta_i(k)}x^{(i)}(k-1) +(A^{(i)})^{-1} \left(e^{A^{(i)}\Delta_i(k)}-I\right)B(k)u_i(k),\\
y^{(i)}(k)&=C(k)x^{(i)}(k),
\end{aligned}
\end{equation*}
where $A^{(i)}\in\mathbb{R}^{n\times n}$ is diagonal, while the input-to-state map $B(k)$, the readout $C(k)$, and the positive discretization step $\Delta_i(k)$ depend on the current input. In
particular, we consider
\begin{equation*}
\begin{aligned}
B(k)&=W_Bu(k),\\
C(k)&=\left(W_Cu(k)\right)^\top,\\
\Delta(k)&=\operatorname{softplus}
\left(W_\Delta u(k)+b_\Delta\right),
\end{aligned}
\end{equation*}
where $W_B,W_C\in\mathbb{R}^{n\times D}$, $W_\Delta\in\mathbb{R}^{D\times D}$, and $b_\Delta\in\mathbb{R}^D$ are learnable parameters. The vector $\Delta(k)\in\mathbb{R}^D$ collects the $D$ feature-specific discretization steps.\footnote{In Mamba, the input-dependent step sizes are computed through a low-rank projection with a learnable bias. Here, we allow a full $D\times D$ projection, which includes the low-rank parameterization as a special case.}

The defining feature of S6 is input-dependent selectivity: the current input determines the discretization step, which controls how the state is updated. However, this dependence enters through the exponential discretization, while $B(k)$ and $C(k)$ introduce additional input-dependent operations. To isolate the role of gating, we simplify these components separately.

First, using the first-order expansions
\begin{equation*}
\begin{aligned}
e^{A^{(i)}\Delta_i(k)} &=I+\Delta_i(k)A^{(i)} +\mathcal{O}\left(\Delta_i(k)^2\right),\\ (A^{(i)})^{-1} \left(e^{A^{(i)}\Delta_i(k)}-I\right)
&=\Delta_i(k)I+\mathcal{O}\left(\Delta_i(k)^2\right),
\end{aligned}
\end{equation*}
we obtain the approximate recurrence
\begin{equation*}
\begin{aligned}
x^{(i)}(k)\approx &\left(I+\Delta_i(k)A^{(i)}\right)x^{(i)}(k-1)+\Delta_i(k)B(k)u_i(k).
\end{aligned}
\end{equation*}

We next replace the input-dependent maps $B(k)$ and $C(k)$ with feature-specific, time-invariant parameters $B^{(i)}\in\mathbb{R}^n$ and
$C^{(i)}\in\mathbb{R}^{1\times n}$. This removes other sources of input dependence, leaving the discretization step as the sole mechanism controlling selectivity. The resulting~simplified~model~is
\begin{equation}
\label{eq:S6-IG}
\begin{aligned}
x^{(i)}(k) &=\left(I+\Delta_i(k)A^{(i)}\right)x^{(i)}(k-1) +\Delta_i(k)B^{(i)}u_i(k),\\
y^{(i)}(k)&=C^{(i)}x^{(i)}(k).
\end{aligned}
\end{equation}

Finally, we replace the scalar step $\Delta_i(k)$ with the component-wise gate
\begin{equation*}
\Delta_{\mathrm{IG}}^{(i)}(k) =\operatorname{diag}\left(\sigma\left(w_u^{(i)}u_i(k)+b^{(i)}\right)\right),
\end{equation*}
where $w_u^{(i)},b^{(i)}\in\mathbb{R}^n$ and $\sigma(\cdot)$ is applied component-wise. As in S6, the gate is computed from an affine function of the input followed by a pointwise nonlinearity. The main difference is that each state coordinate now has its own gating coefficient, determined by the corresponding input feature alone. We use a sigmoid instead of a softplus to ensure that the gating coefficients lie in $(0,1)$. Together with suitable constraints on $A^{(i)}$, this allows us to enforce positive and stable state-transition coefficients, as required by the models in Section~\ref{sec:COFFEE}. Restricting the gate to depend on $u_i(k)$ rather than the full input vector also reduces the number of gating weights from $D^2$ in the full-projection representation above to $nD$. The corresponding bias vectors introduce $D$ and $nD$ additional parameters, respectively.

With this choice, \eqref{eq:S6-IG} becomes
\begin{equation}
\label{eq:S6-IG2}
\begin{aligned}
x^{(i)}(k)&=\left(I+\Delta_{\mathrm{IG}}^{(i)}(k)A^{(i)}\right)x^{(i)}(k-1)+\Delta_{\mathrm{IG}}^{(i)}(k)B^{(i)}u_i(k),\\
y^{(i)}(k)&=C^{(i)}x^{(i)}(k).
\end{aligned}
\end{equation}
Thus, IG retains the input-dependent gating principle of S6 while adopting a simpler discretization and a component-wise gating structure.

\subsection{Normalization of the input-to-state vector}
\label{app:redundancy}

We now show that, under a mild nondegeneracy assumption, the input-to-state vector $B^{(i)}$ in \eqref{eq:S6-IG2} can be fixed to $\mathbf{1}_n$ without changing the represented input-output maps.

Assume that every entry of $B^{(i)}$ is nonzero, and define the invertible diagonal matrix
\begin{equation*}
T_i=\operatorname{diag}\left(B_1^{(i)},\ldots,B_n^{(i)}\right).
\end{equation*}
By construction, $T_i^{-1}B^{(i)}=\mathbf{1}_n$. Consider the change of coordinates
\begin{equation*}
x^{(i)}(k)=T_i z^{(i)}(k).
\end{equation*}
Since $A^{(i)}$,
$\Delta_{\mathrm{IG}}^{(i)}(k)$, and $T_i$
are diagonal, they commute, and hence
\begin{equation*}
T_i^{-1}\Delta_{\mathrm{IG}}^{(i)}(k)A^{(i)}T_i=\Delta_{\mathrm{IG}}^{(i)}(k)A^{(i)}.
\end{equation*}
Moreover,
\begin{equation*}
T_i^{-1}\Delta_{\mathrm{IG}}^{(i)}(k)B^{(i)}=\Delta_{\mathrm{IG}}^{(i)}(k)\mathbf{1}_n.
\end{equation*}
The transformed recurrence is therefore
\begin{equation*}
\begin{aligned}
z^{(i)}(k)&=\left(I+\Delta_{\mathrm{IG}}^{(i)}(k)A^{(i)}\right) z^{(i)}(k-1)+\Delta_{\mathrm{IG}}^{(i)}(k)
\mathbf{1}_n u_i(k),
\end{aligned}
\end{equation*}
with output
\begin{equation*}
y^{(i)}(k)=\widetilde C^{(i)}z^{(i)}(k),\qquad\widetilde C^{(i)}=C^{(i)}T_i.
\end{equation*}
The initial condition transforms as $z^{(i)}(0)=T_i^{-1}x^{(i)}(0)$. Thus, every IG model with a nonzero input-to-state vector $B^{(i)}$ has an equivalent representation with $B^{(i)}=\mathbf{1}_n$. Relabeling $z^{(i)}$ as $x^{(i)}$ and $\widetilde C^{(i)}$ as $C^{(i)}$ gives the normalized recurrence introduced in Section~\ref{sec:COFFEE}.

The same argument extends to the SG and DG models, including their learnable gate biases. Indeed, under the transformation $x^{(i)}(k)=T_i z^{(i)}(k)$, the state-dependent gating term becomes
\begin{equation*}
w_x^{(i)}\odot x^{(i)}(k-1) =\widetilde w_x^{(i)}\odot z^{(i)}(k-1),
\end{equation*}
where $\widetilde w_x^{(i)}=B^{(i)}\odot w_x^{(i)}$. 
Consequently, the SG gate can be written as
\begin{equation*}
\Delta_{\mathrm{SG}}^{(i)}(k)=\operatorname{diag}\left(\sigma\left(\widetilde w_x^{(i)}\odot z^{(i)}(k-1)+b^{(i)}\right)\right),
\end{equation*}
while the DG gate takes the form
\begin{equation*}
\Delta_{\mathrm{DG}}^{(i)}(k)=\operatorname{diag}\left(\sigma\left(w_u^{(i)}u_i(k)+\widetilde w_x^{(i)}\odot z^{(i)}(k-1)+b^{(i)}\right)\right).
\end{equation*}
Both gates retain their original functional forms: only the state-gating weights are reparameterized, while the input-gating weights and biases remain unchanged. Therefore, fixing
$B^{(i)}=\mathbf{1}_n$ entails no loss of expressivity for any of the three model classes under the stated nondegeneracy assumption.

\section{Proofs of the expressivity separation}
\label{app:expressivity_proofs}

\subsection{Proof of Theorem~\ref{thm:thm1}}
\label{app:proof_thm1}

Fix an arbitrary vocabulary size $d_V\geq 2$ and sequence length $L\geq 3$. We consider the SG model with $D=n=1$, $A^{(1)}=-1$, $w_x^{(1)}=1$, $b^{(1)}=0$, and readout $y(k)=x(k)$. The resulting recurrence is
\begin{equation}
\label{eq:sg_proof}
x(k)=x(k-1)+\sigma(x(k-1))
\bigl(u(k)-x(k-1)\bigr),
\end{equation}
where $\sigma(x)=1/(1+e^{-x})$.

For a parameter $R>0$ to be chosen sufficiently large, set
\begin{equation}
\label{eq:sg_embeddings}
x(0)=-R,\qquad
u_{\rm tr}=e^{3R/2},\qquad
u_j=-2R-j,\quad j=2,\ldots,d_V.
\end{equation}
We show that, as $R\to\infty$, the final state approaches the target embedding uniformly over all admissible sequences. This will imply that a sufficiently large but finite choice of $R$ guarantees perfect nearest-neighbor decoding.

Throughout the proof, asymptotic notation refers to the limit $R\to\infty$, with $d_V$ and $L$ fixed. In particular, $f(R)=o(1)$ means that $f(R)\to0$ as $R\to\infty$.

Let $\bar t\in\{1,\ldots,L-2\}$ denote the position of the first trigger, and let $u_j=u(\bar t+1)$ be the corresponding target.

\paragraph{Before the first trigger.}
By \eqref{eq:sg_embeddings}, the initial state is $x(0)=-R$, while all inputs preceding the first trigger belong to $[-2R-d_V,-2R-2]$. Since $0<\sigma(x)<1$, the recurrence \eqref{eq:sg_proof} expresses $x(k)$ as a convex combination of $x(k-1)$ and $u(k)$. Thus, by induction, the state remains in the interval $[-2R-d_V,-R]$, i.e.,
$$
-2R-d_V\leq x(k)\leq -R, \qquad 0\leq k<\bar t.
$$
For $1\leq k<\bar t$, the bound $x(k-1)\leq -R$ and the inequality $\sigma(x)\leq e^x$ imply that $\sigma(x(k-1))\leq e^{-R}$. Moreover, both $u(k)$ and $x(k-1)$ belong to $[-2R-d_V,-R]$, so that $|u(k)-x(k-1)|\leq R+d_V$. It follows from \eqref{eq:sg_proof} that
$$
|x(k)-x(k-1)| =\sigma(x(k-1))|u(k)-x(k-1)| \leq (R+d_V)e^{-R}.
$$
Summing the state increments from time $0$ to $k$
and using $x(0)=-R$, the triangle inequality gives
$$
\begin{aligned}
|x(k)-x(0)|=|x(k)+R| =\left|\sum_{t=1}^{k} \bigl(x(t)-x(t-1)\bigr)\right| &\leq \sum_{t=1}^{k}|x(t)-x(t-1)|\\
&\leq k(R+d_V)e^{-R}\\
& \leq L(R+d_V)e^{-R}=o(1).
\end{aligned}
$$
Since $k<L$ and $d_V,L$ are fixed, the bound tends to zero as $R\to\infty$, independently of the input sequence and the position of the first trigger. In particular,
\begin{equation}
\label{eq:before_first_trigger}
x(\bar t-1)=-R+o(1).
\end{equation}

\paragraph{At the first trigger.}
By \eqref{eq:before_first_trigger} and $\sigma(x)=e^x/(1+e^x)$, we have
$$
\sigma(x(\bar t-1)) =\frac{e^{-R+o(1)}}{1+e^{-R+o(1)}} =e^{-R}(1+o(1)).
$$
Thus, when the trigger $u(\bar t)=e^{3R/2}$ arrives, \eqref{eq:sg_proof} gives
$$
x(\bar t) =x(\bar t-1)+\sigma(x(\bar t-1)) \bigl(e^{3R/2}-x(\bar t-1)\bigr) =e^{R/2}(1+o(1)).
$$
In particular, $x(\bar t)\to+\infty$, and for all sufficiently large $R$,
\begin{equation}
\label{eq:after_first_trigger}
\frac12 e^{R/2}\leq x(\bar t)\leq 2e^{R/2}.
\end{equation}

\paragraph{Storing the target.}
At time $k=\bar t+1$, immediately after the first trigger, the input is the target $u(\bar t+1)=u_j=-2R-j$. By \eqref{eq:sg_proof},
$$
x(\bar t+1)-u_j =\bigl(1-\sigma(x(\bar t))\bigr) \bigl(x(\bar t)-u_j\bigr).
$$
For sufficiently large $R$, we have $x(\bar t)>0>u_j$ by \eqref{eq:after_first_trigger}, so that $x(\bar t+1)-u_j\geq 0$. Using $1-\sigma(x)\leq e^{-x}$, the bounds in \eqref{eq:after_first_trigger}, and $j\leq d_V$, we obtain
$$
\begin{aligned}
0\leq x(\bar t+1)-u_j &\leq e^{-x(\bar t)} \bigl(x(\bar t)+2R+j\bigr)\\
&\leq e^{-\frac12 e^{R/2}}\bigl(2e^{R/2}+2R+d_V\bigr) =:\varepsilon_R.
\end{aligned}
$$
Since $\varepsilon_R\to 0$ as $R\to\infty$, the target is stored with a vanishing error, namely $x(\bar t+1)\in[u_j,u_j+\varepsilon_R]$.

\paragraph{Preserving the target.}
We now show that this small storage error remains negligible until the final trigger. For $\bar t+2\leq k\leq L-1$, all inputs are non-trigger tokens and hence, by \eqref{eq:sg_embeddings}, belong to $[-2R-d_V,-2R-2]$.

Since $u_j=-2R-j$, with $2\leq j\leq d_V$, the bound $0\leq x(\bar t+1)-u_j\leq\varepsilon_R$ established above implies that
$$
x(\bar t+1)\in I_R:= [-2R-d_V,-2R-2+\varepsilon_R].
$$
All subsequent inputs before the final trigger also belong to $I_R$. Since the recurrence \eqref{eq:sg_proof} expresses each new state as a convex combination of the previous state and the current input, induction yields
$$
x(k)\in I_R,\qquad \bar t+1\leq k\leq L-1.
$$
Therefore, for each $\bar t+2\leq k\leq L-1$, both $x(k-1)$ and $u(k)$ belong to $I_R$. Using the upper endpoint of this interval and the inequality $\sigma(x)\leq e^x$, we obtain
$$
\sigma(x(k-1)) \leq e^{x(k-1)} \leq e^{-2R-2+\varepsilon_R}.
$$
Moreover, since both $u(k)$ and $x(k-1)$ belong to $I_R$, whose length is at most $d_V+\varepsilon_R$, we have
$$
|u(k)-x(k-1)|\leq d_V+\varepsilon_R.
$$
Using the bounds above in \eqref{eq:sg_proof} gives
$$
|x(k)-x(k-1)|=\sigma(x(k-1))|u(k)-x(k-1)|\leq (d_V+\varepsilon_R) e^{-2R-2+\varepsilon_R}.
$$
Summing the state increments from time $\bar t+1$ to $L-1$, and recalling that $|x(\bar t+1)-u_j|\leq\varepsilon_R$, the triangle inequality gives
$$
\begin{aligned}
|x(L-1)-u_j| &\leq |x(\bar t+1)-u_j| +\sum_{k=\bar t+2}^{L-1}|x(k)-x(k-1)|\\
&\leq \varepsilon_R +L(d_V+\varepsilon_R) e^{-2R-2+\varepsilon_R}=o(1).
\end{aligned}
$$
Since $\varepsilon_R\to0$ as $R\to\infty$ and $d_V,L$ are fixed, the right-hand side tends to zero. Therefore,
\begin{equation}
\label{eq:target_preserved}
x(L-1)=u_j+o(1),
\end{equation}
showing that the target is preserved, up to a vanishing error, until the final trigger.

\paragraph{At the final trigger.}
By \eqref{eq:target_preserved}, we have $x(L-1)=u_j+o(1)=-2R-j+o(1)$. At time $k=L$, the input is again the trigger, $u(L)=e^{3R/2}$. Using \eqref{eq:sg_proof} and $\sigma(x)\leq e^x$, we obtain
$$
\begin{aligned}
|x(L)-x(L-1)| &=\sigma(x(L-1)) \bigl|e^{3R/2}-x(L-1)\bigr|\\ 
&\leq e^{-2R-j+o(1)}\bigl(e^{3R/2}+2R+j+o(1)\bigr)=o(1).
\end{aligned}
$$
Combining this estimate with \eqref{eq:target_preserved}, the triangle inequality yields
$$
|x(L)-u_j| \leq |x(L)-x(L-1)| +|x(L-1)-u_j|=o(1).
$$
Since all the preceding bounds are uniform over admissible input sequences for fixed $d_V$ and $L$, we can choose a sufficiently large, finite $R$ such that
$$
|x(L)-u_j|<\frac12
$$
for every admissible sequence.

To conclude, recall from \eqref{eq:sg_embeddings} that any two distinct token embeddings are separated by at least one. Hence, for any
other token embedding $u_i\neq u_j$, the reverse triangle inequality yields
$$
|x(L)-u_i|\geq |u_i-u_j|-|x(L)-u_j|\geq 1-|x(L)-u_j| >\frac12>|x(L)-u_j|.
$$
Therefore, $u_j$ is the unique nearest embedding to $x(L)$, and nearest-neighbor decoding recovers the correct target for every admissible sequence. This completes the proof.
\qed

\begin{remark}[Size of the construction]
The error bounds in the proof imply that $R=O(\log(Ld_V))$ suffices to guarantee perfect decoding. Consequently, the trigger embedding $u_{\rm tr}=e^{3R/2}$ grows only polynomially in $L$ and $d_V$.
\end{remark}

\subsection{Proof of Theorem~\ref{thm:thm2}}
\label{app:proof_thm2}


We prove that scalar input-gated models cannot solve a finite induction-head instance. Consider the general scalar input-gated recurrence
$$
x(k)=a(u(k))x(k-1)+b(u(k)), \quad x(0)=x_0,
\qquad
-1 < a(u(k))<1.
$$
For a fixed token $u_i$, write
$$
a_i=a(u_i),
\qquad
b_i=b(u_i),
$$
and define
$$
f_i(x)=a_i x+b_i.
$$

Consider the vocabulary $\{u_1,u_2,u_3\}$, where
$u_3$ is the trigger and $u_1 < u_2$ are the two targets. For an admissible
sequence $s$ of length five, let $x_s$ denote the state reached after the
final trigger, and let $S_1$ ($S_2$) be the set of admissible sequences whose
target is $u_1$ ($u_2$).

\paragraph{Reduction to an averaging identity.}
Let the readout $y = c(x)$ be a strictly monotonic function of $x$. This readout is followed by threshold decoding which selects a symbol in the vocabulary. 
By monotonicity, $c(\cdot)$ is invertible in its codomain so that correct decoding of every sequence in
$S_1 \cup S_2$ requires a threshold $\theta \in \mathbb{R}$ such that either\footnote{Of course, we can arbitrarily select which one of the two inequalities in \eqref{eq:sep}
is strict.}
\begin{equation}\label{eq:sep}
  x_s \leq \theta \ \ \forall s \in S_1
  \quad\text{and}\quad
  x_s > \theta \ \ \forall s \in S_2,
\end{equation}
or the same with both inequalities reversed.
It therefore
suffices to exhibit weights $\alpha_s \ge 0$, not all zero, such that
\begin{equation}\label{eq:cert}
  \sum_{s\in S_1} \alpha_s = \sum_{s\in S_2} \alpha_s
  \qquad\text{and}\qquad
  \sum_{s\in S_1} \alpha_s\, x_s = \sum_{s\in S_2} \alpha_s\, x_s
\end{equation}
for all values of $b_1,b_2,b_3$ and $x_0$. Indeed, if \eqref{eq:sep} held,
multiplying each inequality by $\alpha_s$ and summing would give
\[
  \sum_{s\in S_1} \alpha_s x_s \;\le \; \theta \sum_{s\in S_1} \alpha_s
  \;=\; \theta \sum_{s\in S_2} \alpha_s \;<\; \sum_{s\in S_2} \alpha_s x_s,
\]
contradicting \eqref{eq:cert}; the reversed case is identical.
We construct such weights in three exhaustive cases.

\paragraph{Case 1: $a_1 \in (-1, 0]$, $a_2,a_3 \in (-1, 1)$.}
Let $\beta = f_3(x_0)$. A direct computation gives
\begin{align*}
  f_2\circ f_1\circ f_1(\beta) &= a_1^2 a_2\, \beta + a_2(1+a_1)\, b_1 + b_2,\\
  f_1\circ f_2\circ f_1(\beta) &= a_1^2 a_2\, \beta + (1+a_1 a_2)\, b_1 + a_1\, b_2,\\
  f_1\circ f_1\circ f_2(\beta) &= a_1^2 a_2\, \beta + (1+a_1)\, b_1 + a_1^2\, b_2,
\end{align*}
and comparing coefficients yields
\[
  (-a_1)\, f_2 f_1 f_1(\beta) + (1+a_1)\, f_1 f_2 f_1(\beta) = f_1 f_1 f_2(\beta).
\]
Since the two weights sum to one and $f_3$ is affine, the identity is
preserved by the final trigger:
\[
  (-a_1)\, x_{u_3u_1u_1u_2u_3} + (1+a_1)\, x_{u_3u_1u_2u_1u_3}
  = x_{u_3u_2u_1u_1u_3}.
\]
The first two sequences belong to $S_1$ and the third to $S_2$. For
$a_1 \in (-1,0]$ both weights are nonnegative and they sum to one, so
\eqref{eq:cert} holds (notice that we are only considering three sequences but this is consistent because the weights $\alpha_s$ associated with sequences that we are not considering are selected to be equal to $0$ so that all the sequences in $S_1$ and $S_2$ are formally present as prescribed by \eqref{eq:cert}). In words: the final state of a $u_2$-target sequence
is a convex combination of the final states of two $u_1$-target sequences,
so no~threshold~can~separate~them.

\paragraph{Case 2: $a_2 \in (-1, 0]$, $a_1,a_3 \in (-1, 1)$.}
Exchanging the roles of $u_1$ and $u_2$ in Case 1 gives
\[
  (-a_2)\, x_{u_3u_2u_2u_1u_3} + (1+a_2)\, x_{u_3u_2u_1u_2u_3}
  = x_{u_3u_1u_2u_2u_3},
\]
with the first two sequences in $S_2$ and the third in $S_1$, and
nonnegative weights summing to one for $a_2 \in (-1,0]$.

\paragraph{Case 3: $a_1, a_2 \in (0,1)$, $a_3\in(-1,1)$.}
Consider the sequences
\begin{align*}
  g_1 &:= x_{u_3u_1u_2u_2u_3}, &
  j_1 &:= x_{u_2u_2u_3u_1u_3}, &
  l_1 &:= x_{u_2u_1u_3u_1u_3}, \\
  g_2 &:= x_{u_3u_2u_1u_1u_3}, &
  m_2 &:= x_{u_2u_1u_3u_2u_3}, &&
\end{align*}
where $g_1, j_1, l_1$ have target $u_1$ and $g_2, m_2$ have target $u_2$.
Define
\[
  \lambda := a_1\bigl(1 + a_1 - a_1 a_3 (1 + a_2)\bigr), \quad
  \mu := a_1 a_2 (1 - a_1 a_2), \quad
  \nu := a_2 (1 + a_2)(1 - a_1 a_3).
\]
We claim that
\begin{equation}\label{eq:case3}
  \lambda\, g_1 + \mu\, j_1 + \nu\, l_1 = (\lambda + \mu)\, m_2 + \nu\, g_2,
\end{equation}
which is of the form \eqref{eq:cert}, with total weight
$\lambda + \mu + \nu$ on both sides.

\emph{Removing the final trigger.}
Every sequence ends with the trigger, so $x_s = f_3(\bar x_s)$, where
$\bar x_s$ denotes the state before the last token. We let $\bar g_1,\bar j_1,\bar l_1,\bar g_2,\bar m_2$ denote the states reached after the first four tokens of the sequences defining $g_1,j_1,l_1,g_2,m_2$, respectively. Since both sides of \eqref{eq:case3} carry the same total weight and $f_3$ is affine, it suffices to prove

\begin{equation}
\label{eq:case3_bar}
\lambda\bar g_1+\mu\bar j_1+\nu\bar l_1 =(\lambda+\mu)\bar m_2+\nu\bar g_2.
\end{equation}
These states are given by
\begin{align*}
  \bar g_1 &= a_1a_2^2a_3\,x_0 + a_2^2\,b_1 + (1+a_2)\,b_2 + a_1a_2^2\,b_3,\\
  \bar j_1 &= a_1a_2^2a_3\,x_0 + b_1 + a_1a_3(1+a_2)\,b_2 + a_1\,b_3,\\
  \bar l_1 &= a_1^2a_2a_3\,x_0 + (1+a_1a_3)\,b_1 + a_1^2a_3\,b_2 + a_1\,b_3,\\
  \bar g_2 &= a_1^2a_2a_3\,x_0 + (1+a_1)\,b_1 + a_1^2\,b_2 + a_1^2a_2\,b_3,\\
  \bar m_2 &= a_1a_2^2a_3\,x_0 + a_2a_3\,b_1 + (1+a_1a_2a_3)\,b_2 + a_2\,b_3.
\end{align*}

\emph{Coefficient of $x_0$.}
This coefficient is the product of the slopes of the tokens read, so it
depends only on which tokens occur and not on their order. In \eqref{eq:case3_bar}, $\bar g_1,\bar j_1,\bar m_2$ all have coefficient $a_1a_2^2a_3$, while $\bar l_1,\bar g_2$ have coefficient $a_1^2a_2a_3$. Hence the coefficients of $x_0$ agree.

\emph{Coefficients of $b_1, b_2, b_3$.}
Comparing these coefficients in \eqref{eq:case3_bar} and rearranging, it remains to check
\begin{align}
  &\text{coefficients of $b_1$:} \quad \lambda a_2(a_2-a_3) + \mu(1-a_2a_3) = \nu\, a_1(1-a_3), \label{eq:c3b1}\\
  &\text{coefficients of $b_2$:} \quad (1-a_1a_3)(\lambda a_2 - \mu) = \nu\, a_1^2(1-a_3), \label{eq:c3b2}\\
  &\text{coefficients of $b_3$:} \quad \lambda a_2(1-a_1a_2) - \mu(a_1-a_2) = \nu\, a_1(1-a_1a_2). \label{eq:c3b3}
\end{align}
All three follow from the single quantity
\[
  K := \lambda a_2 - \mu
     = a_1a_2\bigl[(1 + a_1 - a_1a_3 - a_1a_2a_3) - (1 - a_1a_2)\bigr]
     = a_1^2 a_2 (1+a_2)(1-a_3).
\]
Recall $\nu = a_2(1+a_2)(1-a_1a_3)$. For \eqref{eq:c3b2},
$(1-a_1a_3)K = a_1^2a_2(1+a_2)(1-a_3)(1-a_1a_3) = \nu\,a_1^2(1-a_3)$.
For \eqref{eq:c3b1}, substituting $\mu = \lambda a_2 - K$,
\begin{align*}
  \lambda a_2(a_2-a_3) + (\lambda a_2 - K)(1-a_2a_3)
  &= \lambda a_2(1+a_2)(1-a_3) - K(1-a_2a_3)\\
  &= a_2(1+a_2)(1-a_3)\bigl[\lambda - a_1^2(1-a_2a_3)\bigr]\\
  &= a_2(1+a_2)(1-a_3)\, a_1(1-a_1a_3) = \nu\, a_1(1-a_3),
\end{align*}
since $\lambda - a_1^2(1-a_2a_3) = a_1 - a_1^2 a_3$.
For \eqref{eq:c3b3}, again with $\mu = \lambda a_2 - K$,
\begin{align*}
  \lambda a_2(1-a_1a_2) - (\lambda a_2 - K)(a_1-a_2)
  &= \lambda a_2(1+a_2)(1-a_1) + K(a_1-a_2)\\
  &= a_2(1+a_2)\bigl[\lambda(1-a_1) + a_1^2(1-a_3)(a_1-a_2)\bigr]\\
  &= a_2(1+a_2)\, a_1(1-a_1a_2)(1-a_1a_3) = \nu\, a_1(1-a_1a_2),
\end{align*}
where the bracket expands to
$a_1(1 - a_1a_2 - a_1a_3 + a_1^2a_2a_3) = a_1(1-a_1a_2)(1-a_1a_3)$.
This proves \eqref{eq:case3_bar} and hence \eqref{eq:case3}.

\emph{Positivity.}
Since $a_1, a_2 \in (0,1)$ and $a_3 < 1$, we have $\mu > 0$ and
$\nu > 0$ (because $a_1 a_3 < 1$), and
\[
  1 + a_1 - a_1 a_3 (1+a_2) \;\ge\; 1 + a_1 - a_1(1+a_2) \;=\; 1 - a_1 a_2 \;>\; 0,
\]
which implies $\lambda > 0$.

\paragraph{Conclusion.}
Every $(a_1,a_2,a_3)\in(-1,1)^3$ falls into at least one of the three cases. Hence, for every admissible choice of parameters, the final states cannot be separated by a threshold. Consequently, no strictly monotone scalar readout followed by nearest-neighbor decoding can solve the task with perfect accuracy. \hfill$\square$

The coefficients of the input sequences used in the proof are reported below to facilitate verification of its correctness.

\begin{align*}
x_{u_3u_1u_1u_2u_3} &= a_1^2a_2a_3^2\,x_0 + a_2a_3(1+a_1)\,b_1 + a_3\,b_2 + (1+a_1^2a_2a_3)\,b_3,\\
x_{u_3u_1u_2u_1u_3} &= a_1^2a_2a_3^2\,x_0 + a_3(1+a_1a_2)\,b_1 + a_1a_3\,b_2 + (1+a_1^2a_2a_3)\,b_3,\\
x_{u_3u_2u_1u_1u_3} &= a_1^2a_2a_3^2\,x_0 + a_3(1+a_1)\,b_1 + a_1^2a_3\,b_2 + (1+a_1^2a_2a_3)\,b_3.
\end{align*}

\begin{align*}
x_{u_3u_2u_2u_1u_3} &= a_1a_2^2a_3^2\,x_0 + a_3\,b_1 + a_1a_3(1+a_2)\,b_2 + (1+a_1a_2^2a_3)\,b_3,\\
x_{u_3u_2u_1u_2u_3} &= a_1a_2^2a_3^2\,x_0 + a_2a_3\,b_1 + a_3(1+a_1a_2)\,b_2 + (1+a_1a_2^2a_3)\,b_3,\\
x_{u_3u_1u_2u_2u_3} &= a_1a_2^2a_3^2\,x_0 + a_2^2a_3\,b_1 + a_3(1+a_2)\,b_2 + (1+a_1a_2^2a_3)\,b_3.
\end{align*}

\begin{align*}
g_1 = x_{u_3u_1u_2u_2u_3} &= a_1a_2^2a_3^2\,x_0 + a_2^2a_3\,b_1 + a_3(1+a_2)\,b_2 + (1+a_1a_2^2a_3)\,b_3,\\
j_1 = x_{u_2u_2u_3u_1u_3} &= a_1a_2^2a_3^2\,x_0 + a_3\,b_1 + a_1a_3^2(1+a_2)\,b_2 + (1+a_1a_3)\,b_3,\\
l_1 = x_{u_2u_1u_3u_1u_3} &= a_1^2a_2a_3^2\,x_0 + a_3(1+a_1a_3)\,b_1 + a_1^2a_3^2\,b_2 + (1+a_1a_3)\,b_3,\\
g_2 = x_{u_3u_2u_1u_1u_3} &= a_1^2a_2a_3^2\,x_0 + a_3(1+a_1)\,b_1 + a_1^2a_3\,b_2 + (1+a_1^2a_2a_3)\,b_3,\\
m_2 = x_{u_2u_1u_3u_2u_3} &= a_1a_2^2a_3^2\,x_0 + a_2a_3^2\,b_1 + a_3(1+a_1a_2a_3)\,b_2 + (1+a_2a_3)\,b_3.
\end{align*}

\begin{remark}[Relaxed assumptions]
The proof establishes the impossibility result for a broader range of parameters than assumed in the theorem. In Cases 1 and 2, the averaging identities remain valid when $a_1=-1$ or $a_2=-1$, respectively, regardless of the other coefficients. In Case 3, the positivity of the weights only requires $a_1,a_2\in(0,1)$ and $a_3\leq 1$. Consequently, the result extends to $a_1,a_2\in[-1,1)$ and $a_3\leq 1$, allowing $a_i=-1$ and imposing no lower bound on the trigger coefficient $a_3$.
\end{remark}

\section{Experimental details}
\label{app:exp_details}

\subsection{Software and hardware}
\label{app:sw-hw}

All experiments are implemented in PyTorch~2.7, using Hydra for configuration management. Experiments were run on NVIDIA RTX/A40 GPUs. 
All models, including the nonlinear, state-gated variants, use a naive sequential implementation rather than the parallelized fixed-point solvers discussed in \citep{gonzalez2024scalable}, at the scale of our experiments this is not yet a bottleneck.

\subsection{Reproducibility and initialization}
\label{app:repr}

\paragraph{Reproducibility:}Each configuration is run with three different seeds ($\{1, 2, 3\}$), and we report results as mean $\pm$ standard deviation across runs. 

\paragraph{Initialization of IG, SG and DG parameters:} The gate weights are drawn
from a standard normal distribution, and the gate bias, where present, is
initialized to zero (see also Table \ref{tab:init}). A parameter present in two models is initialized in the same
way in both, so that differences in trained behavior are not attributable to
differences at the starting point.

\begin{table}[H]
\centering
\renewcommand{\arraystretch}{1.3}
\begin{tabular}{@{} l l l l @{}}
\toprule
\textbf{Parameter} & \textbf{Shape} & \textbf{Present in} & \textbf{Initialization} \\
\midrule
$w_u^{(i)}$ & $n$ & IG, DG & $\mathcal{N}(0,1)$, entrywise \\
$w_x^{(i)}$ & $n$ & SG, DG & $\mathcal{N}(0,1)$, entrywise \\
$b^{(i)}$ & $n$ & IG, SG, DG & $0$ \\
\bottomrule \\
\end{tabular}
\caption{Initialization of the gate parameters of the matched family.}
\label{tab:init}
\end{table}


The state transition matrix is parameterized as
$A^{(i)} = -2\sigma(\widetilde{A}^{(i)})$, so that the diagonal entries of
$A^{(i)}$ lie in $(-1,1)$ and hence $-I < I+\Delta^{(i)}(k)A^{(i)} < I$
(entrywise) for every admissible gate, which guarantees 
Schur stability. 
The parameter $\widetilde{A}^{(i)}$ is initialized to zero, so that $A^{(i)} = -\tfrac12 I$ at the start of training.

\paragraph{Initialization of the readout and the embedding:} The readout
$C^{(i)}\in\mathbb{R}^{1\times n}$ is drawn entrywise from a standard normal
distribution, as the gate weights are. The token embedding are initialized according the Johnson and Lindenstrauss theorem \citep{dasgupta2003elementary}.

\paragraph{S6 implementation:} The S6 model is implemented as in \cite{mamba}.\footnote{We used the standard Pytorch implementation that can be found here: \url{https://github.com/state-spaces/mamba/blob/main/mamba_ssm/modules/mamba_simple.py}}
 

\paragraph{Optimization:} All models are trained with AdamW at the
learning rate reported for each experiment, with every other hyperparameter left
at its PyTorch default: $\beta=(0.9,0.999)$, $\varepsilon=10^{-8}$ and weight
decay $0.01$. 
After an initial warmup we apply the \emph{ReduceLROnPlateau} Pytorch's scheduler with \emph{factor = $0.5$}, \emph{patience = $5$} and \emph{min\_lr = $10^{-6}$}. Furthermore early stopping is applied when validation accuracy reaches $100\%$.


\subsection{Core module architecture for induction head and selective copying}
\label{appendix:core_module_architecture}
In this section we provide a description of the core module architecture employed to address the induction head task \citep{dao2023hungry}. The same architecture is used for the selective copying task \citep{mamba}. A detailed account of all steps involved in data processing is  presented. The operations performed by the core module in our setup are as follows:
\begin{enumerate}
    \item The core module receives a sequence or a batch of sequences as input. The input can be regarded as an array of dimension [$B$, $L$], where $B$ denotes the batch size and $L$ the length of each sequence within the batch.\footnote{More generally, we use the notation [$d_1$,..., $d_l$] to indicate an array having $l$ dimensions, where the first dimension is $d_1$ and the $l$-th dimension is $d_l$.}
 
    \item Each symbol is mapped to a real-valued vector, referred to as the  {\em embedding vector}. The dimensionality of the embedding space is a user-defined hyperparameter and can be adjusted according to the specific requirements of the task.
    Formally, the embedding procedure consists in applying the following \textbf{learned} map:
    \[
    \textrm{emb}: V \rightarrow \mathbb{R}^{D}
    \]
    where $V$ is the model's vocabulary while $D$ denotes the dimension of the embedding  space and sometimes is also called  {\em model dimension}. After applying this mapping, we obtain a three-dimensional array with shape [$B$, $L$, $D$], where $D$ denotes the embedding dimensionality. 
   
    \item The resulting sequence of embedding vectors is fed into a state space model.
    
    \item For each input (i.e. the embedding vector of a symbol) the model outputs another vector of the same dimension, specifically a vector in $\mathbb{R}^{D}$.
    After processing all input sequences, the state space model produces a multidimensional array of shape [$B$, $L$, $D$]. For each batch index $b = 0, \ldots, B-1$, this corresponds to an $L \times D$ matrix containing all output vectors associated with the input sequence.

    \item In general, the outputs produced by the state space model do not correspond exactly to the embedding vectors associated with symbols in $V$. However, the ultimate objective is for the model to generate symbols from the set $V$. To achieve this, the first step involves computing the distance between each output vector and every embedding vector corresponding to symbols in $V$. This operation results in a multidimensional array of shape [$B$, $L$, $|V|$].
  
    After having evaluated the distances we have:
    \[\mathrm{Distances} = 
    \begin{bmatrix}
        \begin{bmatrix}
            d_{0,0}^{(0)} & \cdots & d_{0,|V|-1}^{(0)}\\
            \vdots & \ddots  & \vdots \\
            d_{L-1,0}^{(0)} & \cdots & d_{L-1,|V|-1}^{(0)}\\
        \end{bmatrix}\\
        \vdots \\
        \begin{bmatrix}
            d_{0,0}^{(B-1)} & \cdots & d_{0,|V|-1}^{(B-1)}\\
            \vdots & \ddots  & \vdots \\
            d_{L-1,0}^{(B-1)} & \cdots & d_{L-1,|V|-1}^{(B-1)}\\
        \end{bmatrix}
    \end{bmatrix}
    \]
    where $d_{l,v}^{(b)}$ denotes the distance between the output vector at time $l$ of batch $b$ and the embedding vector of the symbol $v \in V$, for every $l = 0, \ldots, L-1$, $b = 0, \ldots, B-1$, and $v \in V$.
    \item We now apply the softmin function along the rows of the matrices within the multidimensional array $\mathrm{Distances}$. This operation produces another multidimensional array, denoted as $\mathrm{Softmin}$, which has the same dimensions as $\mathrm{Distances}$, namely [$B$, $L$, $|V|$].

    By applying the softmin function to the rows of $\mathrm{Distances}$ array, we obtain a probability  distribution where the smallest $d_{l,v}^{(b)}$ for every $l=0,...,L-1$, $b=0,...,B-1$ and $v \in V$ has the maximum probability. 
    \newline
    \newline
    The resulting matrix after this step is:
    \[\mathrm{Softmin} = 
    \begin{bmatrix}
        \begin{bmatrix}
            s_{0,0}^{(0)} & \cdots & s_{0,|V|-1}^{(0)}\\
            \vdots & \ddots  & \vdots \\
            s_{L-1,0}^{(0)} & \cdots & s_{L-1,|V|-1}^{(0)}\\
        \end{bmatrix}\\
        \vdots \\
        \begin{bmatrix}
            s_{0,0}^{(B-1)} & \cdots & s_{0,|V|-1}^{(B-1)}\\
            \vdots & \ddots  & \vdots \\
            s_{L-1,0}^{(B-1)} & \cdots & s_{L-1,|V|-1}^{(B-1)}\\
        \end{bmatrix}
    \end{bmatrix}
    \]
    
    \item After applying the softmin function, we then apply an element-wise transformation known as the  logit function. The logit function maps probabilities from the interval $(0,1)$ to the entire real line $\mathbb{R}$.

    In this manner, we obtain another multidimensional array, which we denote as $\mathrm{Logits}$. Each entry is given by $\mathrm{Logits}_{k,i,j} = \logit(\mathrm{Softmin}_{k,i,j})$. Naturally, the $\mathrm{Logits}$ array has the same dimensions as $\mathrm{Softmin}$, namely [$B$, $L$, $|V|$]. The resulting matrix after this step is:
    \[\mathrm{Logits} = 
    \begin{bmatrix}
        \begin{bmatrix}
            \logit_{0,0}^{(0)} & \cdots & \logit_{0,|V|-1}^{(0)}\\
            \vdots & \ddots  & \vdots \\
            \logit_{L-1,0}^{(0)} & \cdots & \logit_{L-1,|V|-1}^{(0)}\\
        \end{bmatrix}\\
        \vdots \\
        \begin{bmatrix}
            \logit_{0,0}^{(B-1)} & \cdots & \logit_{0,|V|-1}^{(B-1)}\\
            \vdots & \ddots  & \vdots \\
            \logit_{L-1,0}^{(B-1)} & \cdots & \logit_{L-1,|V|-1}^{(B-1)}\\
        \end{bmatrix}
    \end{bmatrix}
    \]
    \item To conclude, the core module architecture produces as output two multidimensional arrays. The first is exactly $\mathrm{Logits}$. The second is a multidimensional array of dimensions [$B$, $L$], obtained by applying the $\argmax$ along the last dimension of $\mathrm{Logits}$. This means that we compute the $\argmax$ over all rows of every matrix within the multidimensional array $\mathrm{Logits}$. The resulting array of shape [$B$, $L$] is denoted as $\mathrm{Predictions}$. Formally, this matrix is constructed as follows:
    \[
    \mathrm{Predictions} = 
    \begin{bmatrix}
        \argmax_{\substack{v \in V}} \quad \logit_{0,v}^{(0)} & \cdots & \argmax_{\substack{v \in V}} \quad \logit_{L-1,v}^{(0)}\\
        & \vdots & \\
        \argmax_{\substack{v \in V}} \quad \logit_{0,v}^{(B-1)} & \cdots & \argmax_{\substack{v \in V}} \quad \logit_{L-1,v}^{(B-1)}
    \end{bmatrix}
    \]
    This matrix, for each row (i.e. for each batch), contains the predicted symbols for all time steps $l = 0, \ldots, L-1$. Notice that, as a consequence of the previous operations, taking the $\argmax$ corresponds to selecting, for every batch and every time step, the symbol whose embedding vector is closest (with respect to the Euclidean norm) to the output produced by the state space model. We use cross-entropy loss as the loss function, which requires the logits as input. Therefore, the $\mathrm{Logits}$ array is used for computing the loss, while the $\mathrm{Predictions}$ array is used to evaluate the accuracy.
\end{enumerate}

\begin{figure}
\begin{center}
\begin{tikzpicture}[
  node distance=1.5cm and 2.5cm,
  block/.style={rectangle, draw, minimum width=3cm, minimum height=1cm, align=center},
  arrow/.style={-{Latex}, thick},
  scale=0.75,every node/.style={scale=0.75},
]

\node[block] (sequence) {Input sequence};
\node[block, below=0.5cm of sequence] (embedding) {Embedding};
\node[block, below=0.5cm of embedding] (ssm) {State space model};
\node[block, below=0.5cm of ssm] (distance) {Distances evaluation};
\node[block, below=0.5cm of distance] (softmin) {softmin applied to rows};
\node[block, below=0.5cm of softmin] (logit) {logit applied entry-wise};
\node[block, below left=1.25cm of logit] (branch1) {Logits};
\node[block, below right=1.25cm of logit] (branch2) {Predictions};

\draw[arrow] (sequence) -- (embedding);
\draw[arrow] (embedding) -- (ssm);
\draw[arrow] (ssm) -- (distance);
\draw[arrow] (distance) -- (softmin);
\draw[arrow] (softmin) -- (logit);
\draw[arrow] (logit) -| node[below, pos=0.5, xshift=30pt] {First output} (branch1.north);
\draw[arrow] (logit) -| node[below, pos=0.5, xshift=-35pt] {Second output} (branch2.north);
\end{tikzpicture}
\end{center}
\caption{Block diagram of the core module architecture used for induction head and selective copying, showing the main processing steps from input embeddings to output predictions.}
\label{fig:coremodule}
\end{figure}
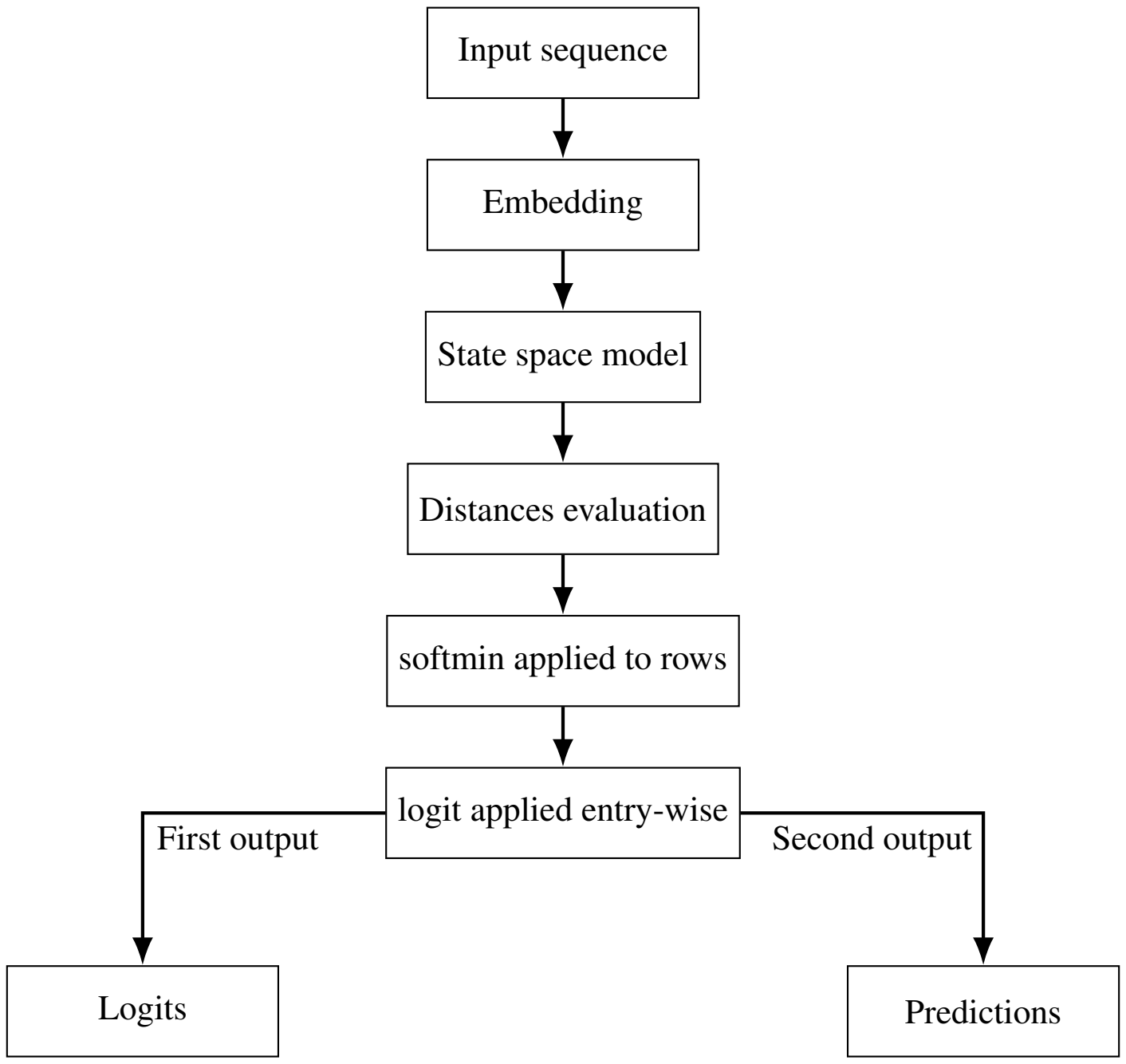

The core module architecture described above is summarized in Figure~\ref{fig:coremodule}.

\subsection{Choice of the base configuration for induction head and selective copying}
\label{app:base-config}

We use an embedding dimension $D=64$ and a state dimension $n=8$ across all models and tasks. The embedding dimension was selected through a preliminary experiment on the induction-head task with $|V|=1024$ and $L_{\mathrm{seq}}=128$. Specifically, we evaluated $D\in\{16,32,64\}$ using SG with $n=8$, learning rate $10^{-2}$, 20 epochs, and a single seed (42). Test accuracy increased with $D$, reaching $6.2\%$, $19.1\%$, and $36.7\%$, respectively, with the loss still decreasing at $D=64$. We therefore selected $D=64$, the largest dimension compatible with our GPU memory constraints. The same configuration is used for all models and tasks to ensure a consistent comparison. 

\subsection{MNIST Architecture}
\label{appendix:mnist_architecture}


For the MNIST experiments, we adopt the architecture introduced in \citep{coffee}. Unlike the standard sequential MNIST setup, which processes individual pixels sequentially, this architecture processes each image as sequences of rows and columns in both forward and reverse directions. This allows us to compare the different gating strategies within a common SSM-based classification architecture. We describe the architecture below for completeness. With the MNIST dataset, we treat an image as a sequence of twenty-eight symbols (i.e. one for each row). For each symbol $s_i$, $i=0,\dots,27$, its embedding vector in $\mathbb{R}^{28}$ is given by the $i$-th row of the image. Thus, processing an image is equivalent to processing a sequence of twenty-eight symbols embedded in $\mathbb{R}^{28}$. To clarify this idea, consider the following diagram.

\begin{center}
\begin{tikzpicture}[scale=0.75,every node/.style={scale=0.75}]
    \draw (-2,0) rectangle (6,3);

    \draw (-2,0.5) -- (6,0.5);
    \draw (-2,2.5) -- (6,2.5);

    \node at (2,2.75) {Embedding of the first symbol in $\mathbb{R}^{28}$};
    \node at (2,1.5) {$\vdots$};
    \node at (2,0.25) {Embedding of the last symbol in $\mathbb{R}^{28}$};

    \draw [->] (6.1,2.75) -- (7,2.75);
    \node[right] at (7,2.75) {First row of the image};
    \draw [->] (6.1,0.25) -- (7,0.25);
    \node[right] at (7,0.25) {Last row of the image};

    \node[left] at (-2,2.75) {$s_0$};
    \node[left] at (-2,0.25) {$s_{27}$};

    \draw[decorate, decoration={brace, amplitude=7.5pt, mirror}] 
        (-2,-0.1) -- (6,-0.1) 
        node[midway, yshift=-20pt]{Overall image};
    \draw[decorate, decoration={brace, amplitude=7.5pt, mirror}] 
        (-2.7,-0.1) -- (-2,-0.1) 
        node[midway, yshift=-20pt]{Symbols};
    
\end{tikzpicture}
\end{center}

\begin{figure}[H]
\begin{center}
\begin{tikzpicture}[
  node distance=1cm and 0.5cm, 
  block/.style={rectangle, draw, minimum width=2cm, minimum height=1cm, align=center},
  arrow/.style={-{Latex}, thick},
  scale=0.75,every node/.style={scale=0.75}
]

\node[block] (input_seq1) {Image by rows};
\node[block, right=of input_seq1] (input_seq2) {Image by cols};
\node[block, right=of input_seq2] (input_seq3) {Image by rows reverse};
\node[block, right=of input_seq3] (input_seq4) {Image by cols reverse};
\node[block, below=0.75cm of input_seq1] (ssm_layer1) {SSM layer};
\node[block, below=0.75cm of input_seq2] (ssm_layer2) {SSM layer};
\node[block, below=0.75cm of input_seq3] (ssm_layer3) {SSM layer};
\node[block, below=0.75cm of input_seq4] (ssm_layer4) {SSM layer};
\node (horizontal_stack) [rectangle, draw, below=1.8cm of $(ssm_layer2)!0.63!(ssm_layer3)$] {Horizontal stack of the last outputs};
\node[block, below=0.75cm of horizontal_stack] (linear_layer1) {Linear layer: $\mathbb{R}^{112} \rightarrow \mathbb{R}^{28}$};
\node[block, below=0.75cm of linear_layer1] (gelu) {GELU};
\node[block, below=0.75cm of gelu] (linear_layer2) {Linear layer: $\mathbb{R}^{28} \rightarrow \mathbb{R}^{10}$};
\node[block, below=0.75cm of linear_layer2] (prediction) {$\argmax$};

\draw[arrow] (input_seq1) -- (ssm_layer1);
\draw[arrow] (input_seq2) -- (ssm_layer2);
\draw[arrow] (input_seq3) -- (ssm_layer3);
\draw[arrow] (input_seq4) -- (ssm_layer4);
\draw[arrow] (ssm_layer1) -- (horizontal_stack);
\draw[arrow] (ssm_layer2) -- (horizontal_stack);
\draw[arrow] (ssm_layer3) -- (horizontal_stack);
\draw[arrow] (ssm_layer4) -- (horizontal_stack);
\draw[arrow] (horizontal_stack) -- (linear_layer1);
\draw[arrow] (linear_layer1) -- (gelu);
\draw[arrow] (gelu) -- (linear_layer2);
\draw[arrow] (linear_layer2) -- (prediction);
\end{tikzpicture}
\caption{MNIST classification architecture. Each image is processed by four SSM layers operating on rows and columns in both forward and reverse directions. Their final outputs are concatenated and passed through two linear layers with a GELU activation to predict the digit class.}
\label{fig:archs_MNIST}
\end{center}
\end{figure}
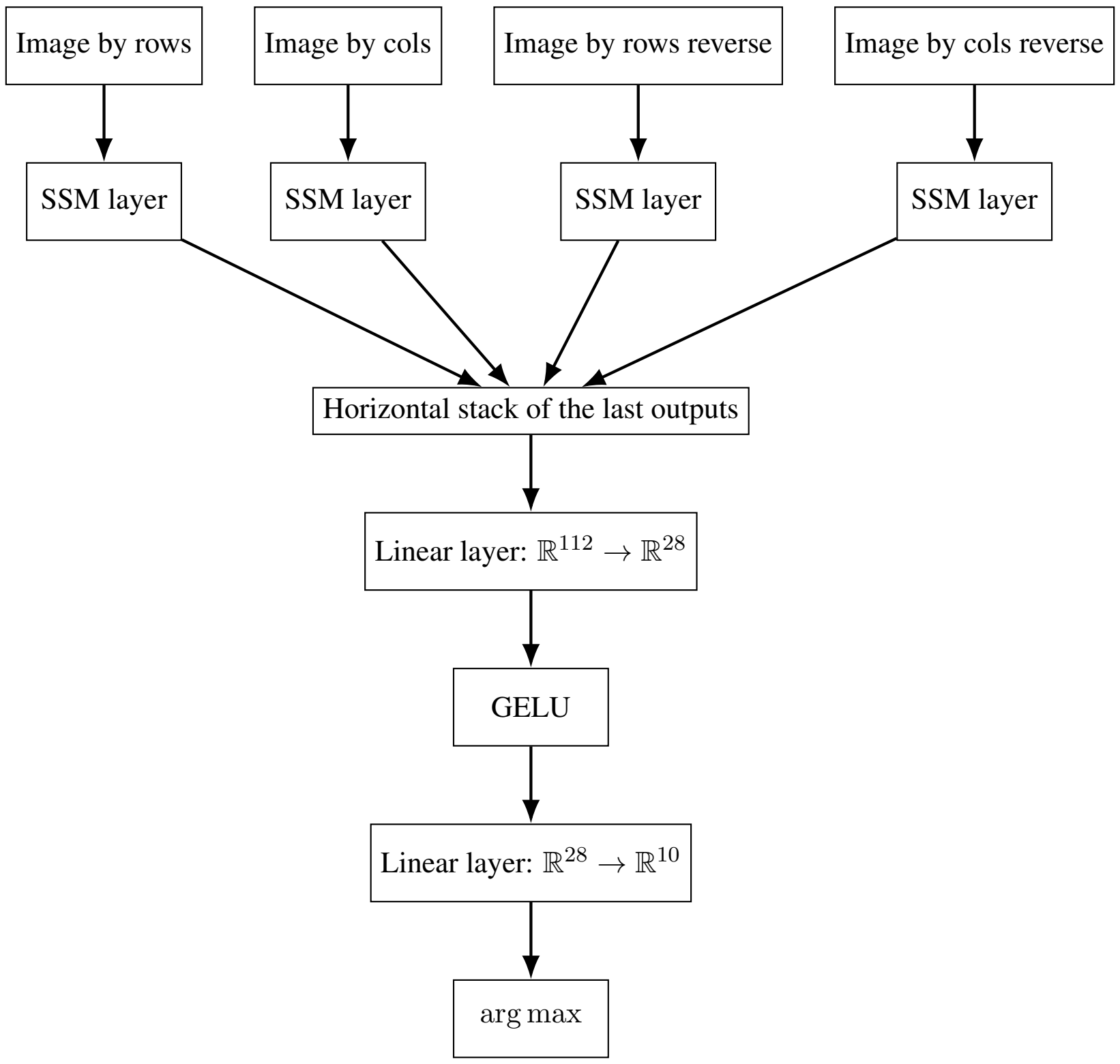

Drawing inspiration from the celebrated two filter formula used for the smoothing problem, see \citep{smoothing_problem_Pavon} and \citep{smoothing_problem_Ferrante}, we can consider the image as a sequence of symbols $s_0 \| \dots \| s_{27}$ (where $\|$ denotes symbols concatenation) , with each $s_i \in \mathbb{R}^{28}$, and build two sequences: one using the rows as embedding vectors and the other using the columns. Let $s_0 \| \dots \| s_{27}$ denote the sequence created from the rows and $s_0^{'} \| \dots \| s_{27}^{'}$ the one from the columns. The index $i=0,\dots,27$, can now be interpreted as time. By thinking of a sequence as a discrete time signal of finite support, given that we have access to all the signals in advance, we process the same image in four ways:
\begin{enumerate}
    \item By rows from $i=0$ to $i=27$;
    \item By columns from $i=0$ to $i=27$;
    \item By rows from $i=27$ to $i=0$;
    \item By columns from $i=27$ to $i=0$.
\end{enumerate}
The resulting architecture is illustrated in Fig.~\ref{fig:archs_MNIST}. In this setup, we process the images by rows, columns, and by rows and columns in reverse order. Each SSM layer produces an output matrix in $\mathbb{R}^{28 \times 28}$, from which we consider only the last row. We then take the last rows of all four matrices and stack them horizontally to obtain a vector in $\mathbb{R}^{112}$. This vector is mapped to the corresponding logits using two linear layers separated by a GELU activation function. The predicted class is again selected as:
\[
\hat{i} = \argmax_{i \in \{0,...,9\}} \quad v_i
\]
where with $\hat{i}$ we denote the predicted class and with $v_i$ we denote the $i$-th entry of the vector $v \in \mathbb{R}^{10}$.

\subsection{Effect of the gate bias on SG training}
\label{app:bias_warm_start}

Table~\ref{tab:ih_added_new} shows SG losing most of its accuracy on induction
head when the gate is given a bias. We wonder whether the degradation arises for expressivity or optimization issues. This appendix reports an
experiment that separates the two possibilities. We take SG models already trained without a bias, copy their weights into new models, and continue training for a further $50$ epochs. Two arms are run, both starting from the same trained weights and using the same data, optimizer and learning rate: a control without a bias, and a model in which the gate is given a bias initialized to zero. The control arm is what makes the comparison readable: without it, a degradation could equally be explained by the act of continuing to train from an already converged point.

\begin{table}[H]
\centering
\renewcommand{\arraystretch}{1.35}
\begin{tabular}{@{} l c c @{}}
\toprule
& \textbf{without bias} & \textbf{with bias} \\
\midrule
From scratch, $50$ epochs & $83.6 \pm 5.0$\% & $32.7 \pm 38.6$\% \\
Warm-started, $50$ further epochs & $97.2\pm 2.4$\% & ${99.4 \pm 0.5}$\% \\
\bottomrule \\
\end{tabular}
\caption{Induction-head test accuracy of SG at
$\mathrm{lr}=3{\cdot}10^{-3}$, trained from a random initialization and
warm-started from an SG model trained without a bias at the same learning rate.
The first row repeats the corresponding entries of
Table~\ref{tab:ih_added_new}.}
\label{tab:bias_warm_start}
\end{table}

The bias is therefore not harmful to the solution, but to the search for it: a random initialization limits the ability of the optimizer to recover an effective solution, while in the neighborhood of a good solution the extra parameter is used productively.

\end{document}